\documentclass[journal]{IEEEtran}
\usepackage{amsmath,graphicx}
\usepackage{color,hyperref}
\usepackage[table,xcdraw]{xcolor}
\newlength\savedwidth
\newcommand\whline{\noalign{\global\savedwidth\arrayrulewidth
		\global\arrayrulewidth 1.25pt}%
	\hline
	\noalign{\global\arrayrulewidth\savedwidth}}

\begin{document}
%\ninept
%
\title{VSSA-NET: Vertical Spatial Sequence Attention Network for Traffic Sign Detection}
\author{Yuan~Yuan,~\IEEEmembership{Senior Member,~IEEE,}
	Zhitong~Xiong,~\IEEEmembership{Student Member,~IEEE,}
	and~Qi~Wang,~\IEEEmembership{Senior~Member,~IEEE}

\IEEEcompsocitemizethanks{
	\IEEEcompsocthanksitem 2019 IEEE. Personal use of this material is permitted. Permission from IEEE must be obtained for all other uses, in any current or future media, including reprinting/republishing this material for advertising or promotional purposes, creating new collective works, for resale or redistribution to servers or lists, or reuse of any copyrighted component of this work in other works..
}}

\markboth{IEEE TRANSACTIONS ON IMAGE PROCESSING}%
{Shell \MakeLowercase{\textit{et al.}}: Bare Demo of IEEEtran.cls for IEEE Journals}

\maketitle

\begin{abstract}
	Although traffic sign detection has been studied for years and great progress has been made with the rise of deep learning technique, there are still many problems remaining to be addressed. For complicated real-world traffic scenes, there are two main challenges. Firstly, traffic signs are usually small-size objects, which makes it more difficult to detect than large ones; Secondly, it is hard to distinguish false targets which resemble real traffic signs in complex street scenes without context information. To handle these problems, we propose a novel end-to-end deep learning method for traffic sign detection in complex environments. Our contributions are as follows: 1) We propose a multi-resolution feature fusion network architecture which exploits densely connected deconvolution layers with skip connections, and can learn more effective features for small-size object; 2) We frame the traffic sign detection as a spatial sequence classification and regression task, and propose a vertical spatial sequence attention (VSSA) module to gain more context information for better detection performance. To comprehensively evaluate the proposed method, we do experiments on several traffic sign datasets as well as the general object detection dataset and the results have shown the effectiveness of our proposed method.
\end{abstract}
\section{Introduction}
\label{sec:intro}
Road signs play a vital role in maintaining traffic safety and conveying road information to drivers or pedestrians, so the signs are designed with regular shapes and striking colors. Detecting and recognizing them automatically is an important sub-module of driver assistant systems and autonomous vehicles. Considering its industrial potential, traffic sign recognition (TSR) systems have been heavily studied in recent years.  Many methods obtain good performance on some traffic sign detection datasets, especially the deep learning based approaches. 

For traffic sign detection and recognition in complicated driving scenes, methods which using hand-craft features such as HOG \cite{dalal2005histograms}, SIFT or color and shape prior are not robust enough for distinguishing real signs from fake ones. From our observation, the main reason is that many objects in complicated driving scenes have similar appearance with traffic signs, and the low level image features are not able to represent the subtle differences. Therefore, these methods are with high false alarm rate, and are not robust enough for real-world applications. 

Fortunately, deep convolutional neural network has brought a vast improvement in image classification performance comparing to conventional methods. By using deep learning models, more powerful representation features can be learned automatically. Moreover, object detection task can also benefit from deep learning methods. Great progress has been made in object detection owing to the resurgence of deep convolutional neural networks. By stacking multiple convolution layers, features learned by these deep models are with higher semantic-level. More powerful features result in lower false alarm rate and higher detection performance. Among these deep learning based object detection methods, region proposal based methods like R-CNN \cite{girshick2014rich}, Fast-RCNN \cite{girshick2015fast}, Faster R-CNN \cite{ren2015faster} and other modifications have achieved state-of-the-art results on many object detection benchmarks, such as PASCAL VOC \cite{everingham2010pascal}, KITTI\cite{geiger2012we}, MS COCO\cite{lin2014microsoft} and so forth. 

\begin{figure}
	\centering
	\includegraphics[width=0.45\textwidth]{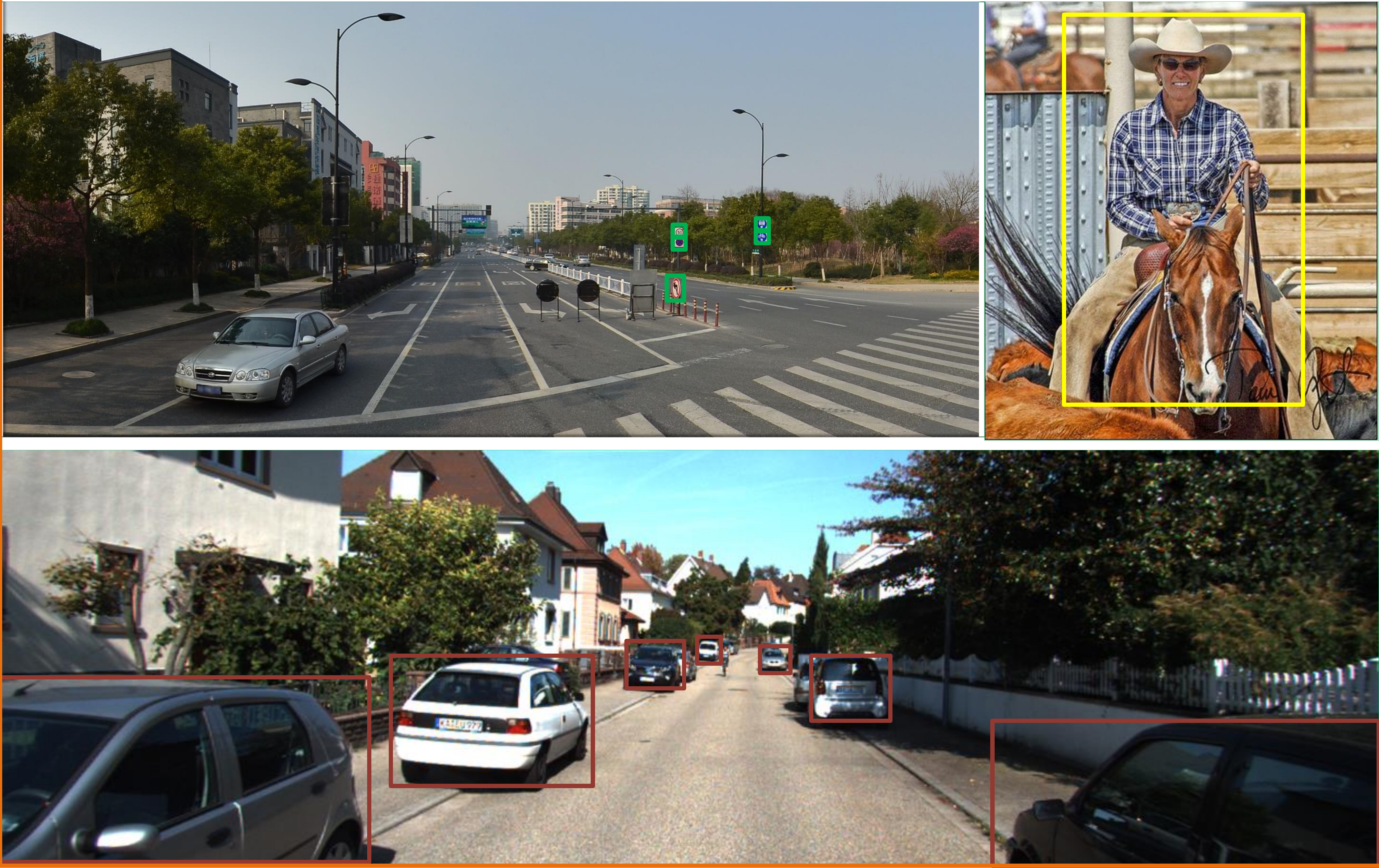}
	\caption{Illustration of the different detection task, including traffic sign detection, person detection and vehicle detection.}
	\label{fig:small}
\end{figure}
\begin{figure}
	\centering
	\includegraphics[width=0.45\textwidth]{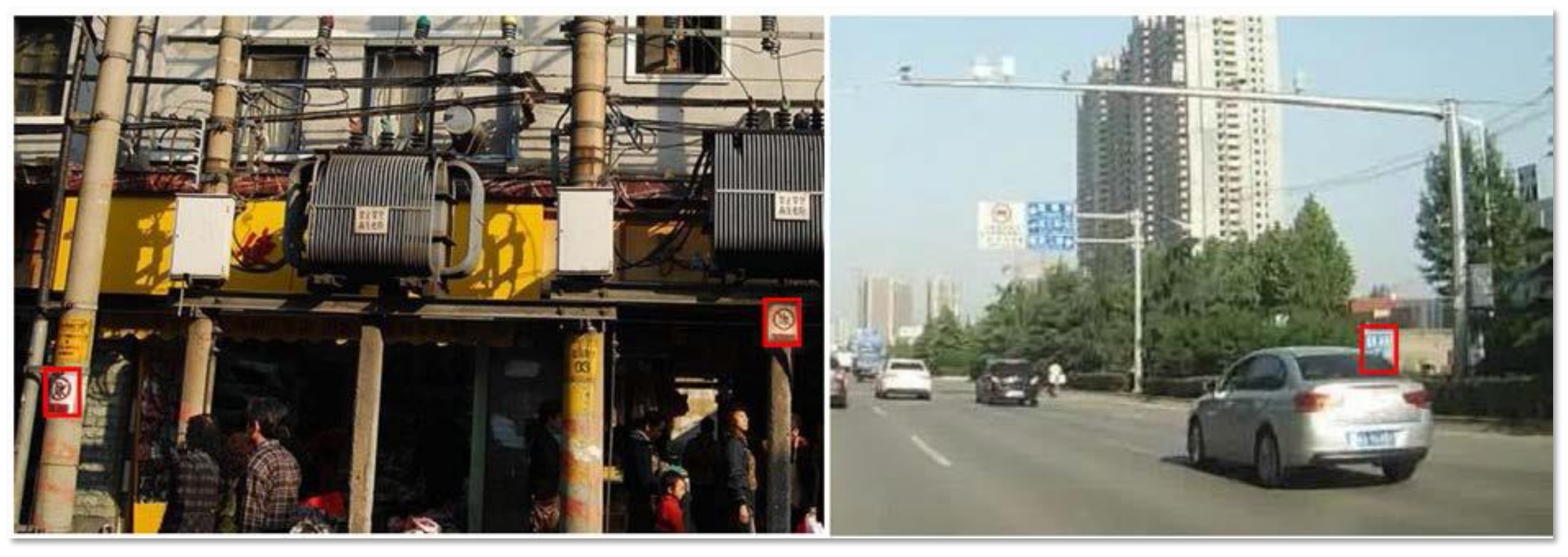}
	\caption{Illustration of the non-traffic signs, which is similar to real traffic signs.}
	\label{fig:complicate}
\end{figure}

\begin{figure*}
	\centering
	\includegraphics[width=0.8\textwidth]{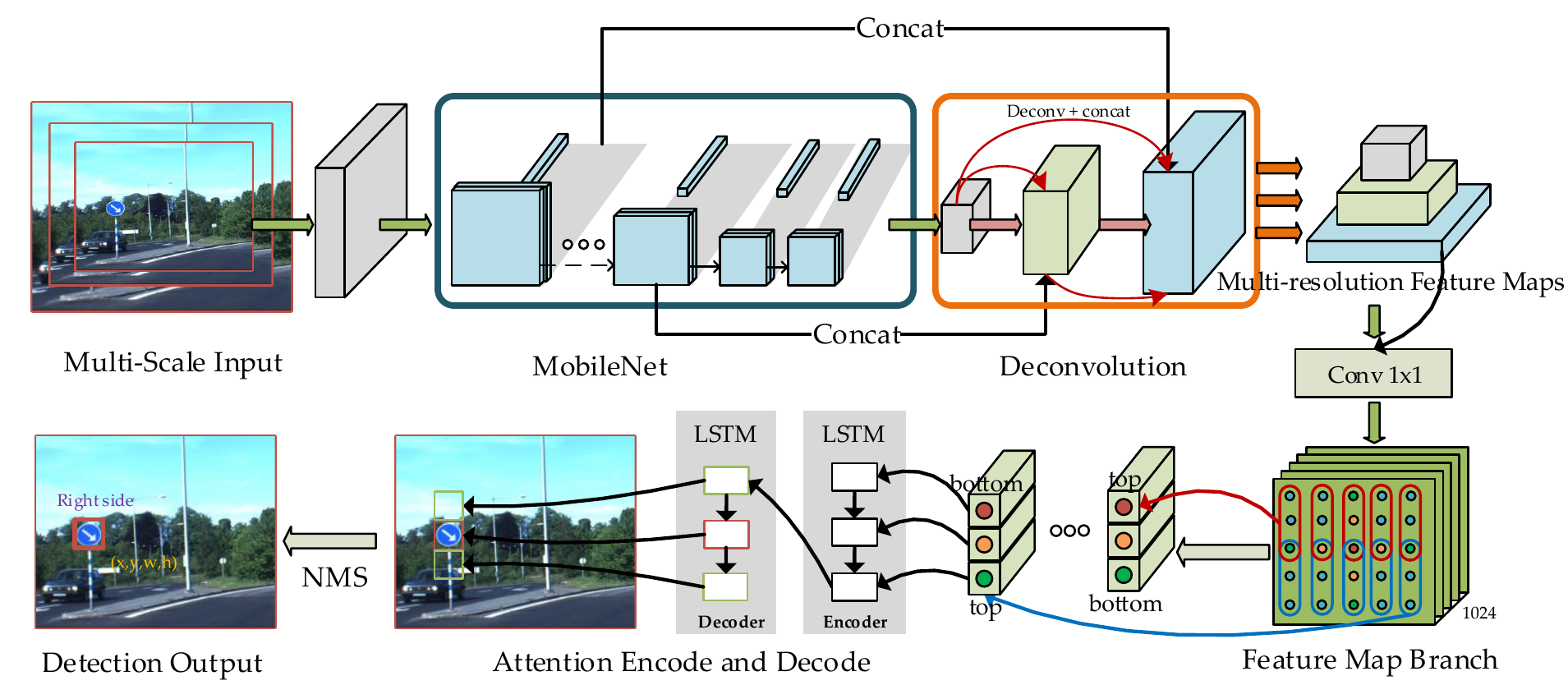}
	\caption{The whole detection framework.}
	\label{fig:network}
\end{figure*}

These deep learning detection methods can be directly applied to detect road signs, as traffic sign detection is a kind of rigid object detection task. However, traffic signs are different from other objects like vehicles, pedestrians in driving scenes. Directly applying general object detection methods cam hardly get satisfactory performance. One reason is that traffic signs have smaller size than many other objects, and usually occupy less than 1\% of each image as shown in Fig.\ref{fig:small}. Moreover, traffic signs are designed to inform drivers to drive in compliance with the traffic rules. Thus for real world applications, traffic signs should be detected as soon as possible to gain more reaction time for drivers. The signs that need to be detected by driver assistant systems are small in size. When the size is small, the appearance information can be used decreases, and it makes the detection more challenging. Another reason is that traffic scene is getting more and more complicated in downtown area. There may exist many billboards or other signs unrelated to traffic, which makes the detection task more challenging. As presented in Fig. \ref{fig:complicate}, the signs not related to traffic may lead to false detections, if no spatial context information is used. Hence, traffic sign detection is a difficult task for real world applications.

As depicted above, the complex traffic scene further makes the detection task difficult in comparison with general object detection. However, humans can exploit context information to recognize small objects in complicated traffic scenes. Taking as example the traffic signs, human can determine whether a target is a sign by the surroundings, especially the vertical direction of the view, such as the pole supporting the sign. Considering that traffic signs locate in specific environments, and some objects co-exist with them, we employ RNN (Recurrent Neural Networks) to explicitly learn the pattern of region sequences rather than merely the local region of interest. By encoding the surrounding visual features into the final feature maps, more discriminative features can be learned to distinguish a true traffic sign from advertising signs. However, directly modeling the region sequences of the image may introduce some irrelevant background, which have no contribution to the traffic sign detection. To address this problem, we design a vertical spatial sequence model with attention mechanism to encode the contextual feature disregarding the noise.

The contributions of this work can be summarized as follows:
\begin{enumerate}
	\item[(1)] We introduce an effective single-shot detector with multi-resolution fused features. To alleviate the small size problem, we employ densely connected deconvolution layers with skip connections to obtain multi-resolution feature maps. We show that well-designed network architecture with dedicated process for small size object can improve the detection performance.
	\item[(2)] We view the traffic sign detection as a region sequence classification and regression task. As aforementioned, traffic signs are small objects occurring in specific patterns. By explicitly modeling the local sequence of regions with attention mechanism, we can gain more context information for better detection accuracy.
	\item[(3)] Considerable experiments has been done to evaluate the performance of the proposed method, including evaluation on two traffic sign detection datasets and one general object detection dataset.
\end{enumerate}

The rest of the paper is organized as follows. Related works about traffic sign detection is shown in Section II. The detail of our method is presented in Section III. The experiments section demonstrated the performance of the proposed method. Finally the conclusion of this paper is given in Section V.

\section{Related Work}

Computer vision methods devised for traffic sign detection can be divided into three categories: color- and shape-based, traditional machine learning based and deep learning based. To make a comprehensive review of the related work, these three kinds of methods will be introduced respectively.

\textbf{Shape and color based methods.} As traffic signs are designed with regular shapes and special colors, many methods try to exploit these strong priors. Such as \cite{gao2008colour} exploits the CIECAM color appearance model and propose to calculate different color attributes including lightness, chroma and hue angle for traffic sign segmentation. The color probability model (CPM) proposed by \cite{yang2016towards} estimates the color distribution of traffic signs from the training samples, and the Ohta space \cite{ohta1980color} is used rather than RGB space. Besides the color-based methods, shape-based detection algorithms are also heavily investigated. Constrained Hough transform is applied in \cite{gonzalez2011automatic} to detect triangular, rectangular, and arrow signs. Other method like Radial symmetry voting \cite{barnes2008real} makes use of the characteristic of traffic sign appearance. While in complicated scenes, color-based and shape-based methods are sensitive to illumination changing, shadows and different weather conditions. Thus color and shape based methods are usually used as the preprocessing or postprocessiong stage of the detection or recognition system.

\textbf{Machine learning based methods.} In recent years, machine learning methods achieves satisfactory results in many computer vision tasks \cite{wang2018hierarchical} \cite{wang2018detecting} \cite{wang2018spectral} including object detection. Traffic sign is a kind of rigid object. Many machine learning based general object detection methods can be applied to detect signs. Viola and Jones \cite{viola2001robust} propose a real-time cascade of boosted Harr-like classifiers detector for face detection, but it is also suitable for traffic sign. Thus many modifications of Viola Jones detector \cite{yuan2016incremental} \cite{chen2008boosted} have improved the performance of traffic sign detection. Another milestone is Aggregated Channel Feature (ACF) detector \cite{dollar2014fast}. ACF detector is compact and effective for rigid object detection in simple scenes. However its limitation is that the shallow feature is not powerful enough in complicated traffic scenes. For pedestrian detection with occlusion under complex scene, a boosted multi-task detector is proposed in \cite{zhu2015boosted} to handle the occlusion problem effectively.

\textbf{Deep learning based methods.} Features extracted by deep learning methods are more semantic. Deep neural networks based detection framework can be divided into two categories: region proposal based methods and single stage based methods. R-CNN \cite{girshick2014rich} first demonstrates the effectiveness of two stage detection frameworks, which uses region proposals and classification subsequently. Other modifications such as Fast R-CNN \cite{girshick2015fast}, Faster R-CNN \cite{ren2015faster}, R-FCN \cite{dai2016r} and Mask R-CNN \cite{he2017mask} have achieved satisfactory performance on public object detection benchmarks. Another kind of deep detection framework is single stage based.  Overfeat \cite{sermanet2013overfeat} first combines object classification neural networks with detection task by sliding windows on the last of shared feature maps. DenseBox \cite{huang2015densebox} employs deconvolution layers to enlarge the feature maps so as to preserve more detail information of images. YOLO \cite{redmon2016you} and SSD \cite{liu2016ssd} are real-time single stage detectors. SSD exploits multi-layer features for detection to improve performance. However, directly using these methods obtains relative poor performance for locating traffic signs in complicated driving scenes. As single-shot detectors are more promising to be real-time, many single shot framework based methods \cite{he2017single} \cite{liao2017textboxes} \cite{liao2018textboxes} are proposed for specific object detection such as scene text detection, which is similar to traffic sign detection in consideration of the complex background.

\textbf{Small-size object detection methods.} To detect objects with small size, many different methods have been proposed. Perceptual Generative Adversarial Networks (GAN) \cite{li2017perceptual} is used to learn super-resolved features in R-CNN for small object detection. MS-CNN \cite{cai2016unified} exploits multiple layer features with different resolutions for multi-scale object detection. For tiny face detection, \cite{hu2017finding} explores three aspects which affect small face detection performance: image resolution, contextual reasoning and scale invariance. In DSSD \cite{fu2017dssd}, deconvolution layers are used to enlarge the feature map resolution for small object detection. The results show that using features with higher resolution can improve small object detection performance. Feature-Fused SSD \cite{Xie2018Feature} explores how to combine features of different layers to boost the detection performance for small objects.

\textbf{Spatial context based methods.}  However, these methods mainly focus on powerful features, the rich context information is relatively neglected. \cite{bell2015inside} propose spatial RNN to encode different direction context information, which improves performance on small-size object. Another method \cite{chen2017spatial} uses spatial memory iterations to encoder object-to-object context.

\begin{figure}
	\centering
	\includegraphics[width=0.4\textwidth]{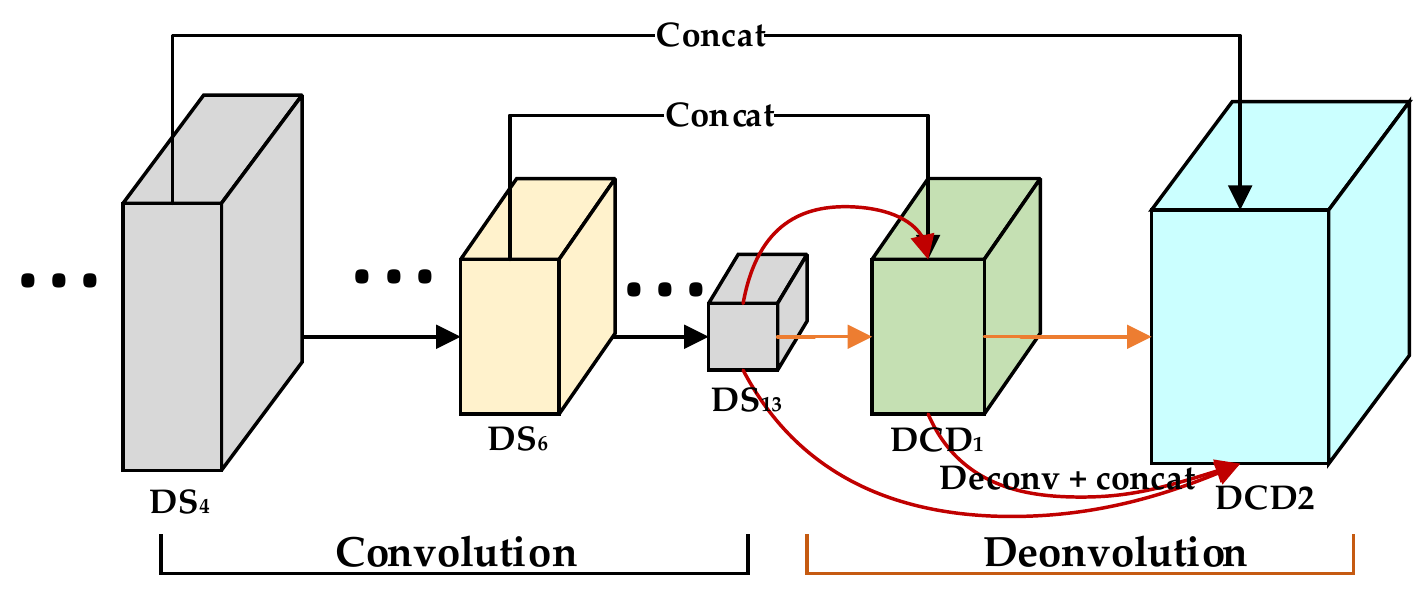}
	\caption{The Multi-Resolution feature learning module.}
	\label{fig:MRF}
\end{figure}

\begin{table*}[]
	\centering
	\caption{VSSA Network Backbone Architecture}
	\label{tabable:arc}
	\begin{tabular}{c|c|c|c|c}
		\hline
		Layer       & Type                                & Channels & Kernel Size & \multicolumn{1}{l}{Stride} \\ \hline
		Conv1       & Convolution                         & 32       & 3 x 3       & 2                          \\ \hline
		DepthConv1  & Depth Separable Convolution x 1     & 64       & 3 x 3       & 1                          \\ \hline
		DepthConv2  & Depth Separable Convolution x 1     & 128      & 3 x 3       & 2                          \\ \hline
		DepthConv3  & Depth Separable Convolution x 1     & 128      & 3 x 3       & 1                          \\ \hline
		DepthConv4  & Depth Separable Convolution x 1     & 256      & 3 x 3       & 2                          \\ \hline
		DepthConv5  & Depth Separable Convolution x 1     & 256      & 3 x 3       & 1                          \\ \hline
		DepthConv6  & Depth Separable Convolution x 1     & 512      & 3 x 3       & 2                          \\ \hline
		DepthConv7  & Depth Separable Convolution x 5     & 512      & 3 x 3       & 1                          \\ \hline
		DepthConv12 & Depth Separable Convolution x 1     & 1024     & 3 x 3       & 1                          \\ \hline
		DepthConv13 & Depth Separable Convolution x 1     & 1024     & 3 x 3       & 2                          \\ \hline
		DCD1        & Densely Connected Deconvolution x 1 & 512      & 3 x 3       & 2                          \\ \hline
		DCD2        & Densely Connected Deconvolution x 1 & 512      & 3 x 3       & 2                          \\ \hline
		DCD3        & Densely Connected Deconvolution x 1 & 512      & 3 x 3       & 2                          \\ \hline
	\end{tabular}
\end{table*}

\section{Our Method}

The whole network is illustrated in Fig. \ref{fig:network}. We adopt MobileNet \cite{howard2017mobilenets} as the backbone for its high time-efficiency. The main process of the system consists of two components. One is the multi-resolution feature learning module, which is used to combine different semantic level features with extra densely connected deconvolution layers. Another module is the vertical spatial sequence attention (VSSA) module, which explicitly encodes the vertical spatial context for more accurate traffic sign classification. During the training stage, an input image is transformed to multi-scales to learn the scale-invariance deep features. Then the multi-resolution feature maps are built through the proposed multi-resolution feature (MRFeature) network. Next, the spatial sequence is encoded along each column of the multi-resolution feature maps by a LSTM network. At last, an attention LSTM decoder layer is exploited to output the classification and detection results simultaneously. The proposed framework is single stage without region proposal process, which makes it fast and can be trained in an end-to-end manner.

\subsection{Multi-Resolution Conv-Deconv Network}
For object detection, higher resolution features are important for providing more accurate location information. Shallower layers of the deep neural network can preserve more object location details. Thus deep detection methods such as SSD\cite{liu2016ssd} and Multi-Scale CNN\cite{cai2016unified} attempt to use the deep features of different layers with multiple resolutions. Another important factor for improving object detection performance is exploiting features with higher semantic level. Inspired by this, we proposed multi-resolution conv-deconv network which takes advantage of densely connected deconvolution layer to magnify the feature maps and obtain higher semantic features simultaneously.

There are two advantages of the conv-deconv feature learning module for detecting traffic signs. One is that deeper convolution networks can extract higher-level semantic features. As de-convolutional layers are usually implemented with the transposed convolution operation, they are essentially the same with convolution layer. These extra learnable layers make the original CNNs deeper by adding several de-convolution layers on top of the original network. Thus the conv-deconv module can learn higher semantic-level representation. Another advantage is that multi-level feature maps with higher resolution are obtained with this module. The proposed conv-deconv architecture fuses low-level and high-level features gradually to enhance the final feature maps with more details, and the higher resolution features can do tremendous help to detecting small objects. 

MobileNet \cite{howard2017mobilenets} proposed depth-wise separable convolution, which factorize the standard convolution into two separable operations. The first one is the depth-wise convolution, which applies a single convolution filter to each input feature map. Then the pointwise convolution combines the depthwise convolution feature maps by using a ${1 \times 1}$ convolution. We add three extra convolution layers to the Mobilenet backbone. The detailed network architecture is shown in Table. \ref{tabable:arc}.

As MobileNet consists of 14 layers in total. We use ${DS_i}$ to present the ${i_{th}}$ depth-separation convolution layer. Then the multi-resolution conv-deconv module can be formulated as follows. The MobileNet outputs ${DS_4}$ and ${DS_6}$ are used as high resolution features to recover the object details in deeper deconvolution layers. Although feature maps with higher-resolution is helpful for detecting small objects, too large feature maps will bring undesirable computational cost. Thus we need to select appropriate layers to balance the computational complexity and the resolution. Considering this, layer ${DS_4}$ and ${DS_6}$ are employed to build the multi-resolution conv-deconv module.

Additionally, concatenating these two layers to the deeper layers can help to alleviate the vanishing-gradient problem. As illustrated in Fig. \ref{fig:MRF}, ${DCD_1}$ and ${DCD_2}$ represent the densely connected deconvolution layers. ${DCD_1}$ layer is computed by 
\begin{equation}
	DC{D_1} = [D{S_6};deconv(D{S_{13}})]. 
\end{equation}

${DCD_2}$ is computed based on ${DCD_1}$. To concatenate layers with lower resolution, we adopt deconvolution operation before the concatenation. This can be formulated as follows:
\begin{equation}
	DC{D_2} = [D{S_4};deconv(DC{D_1});deconv(D{S_{13}})],
\end{equation}
where ${[ \cdot ]}$ denotes the concatenation operation.

Finally ${DS_{13}}$, ${DCD_1}$ and ${DCD_2}$ are taken as the final multi-resolution feature maps for the subsequent sequence modeling. It is worth noting that ${L_2}$ ${normalization}$ is used after the concatenation operation of ${DCD_1}$ and ${DCD_2}$. We modified the final layer of MobileNet from stride 1 to stride 2, so the resolutions of these three feature maps are ${5 \times 5}$, ${10 \times 10}$ and ${19 \times 19}$ respectively.
	
Moreover, scale-invariant representation is significant for computer vision tasks. To detect small objects together with the large ones, in our work we pre-process the input training images to be multi-scale as the supervision to train our network. By sharing parameters with different scale inputs, we make the designed multi-resolution network to learn scale-invariance automatically.

\subsection{Vertical Spatial Sequence Attention Model}
\begin{figure*}
	\centering
	\includegraphics[width=0.9\textwidth]{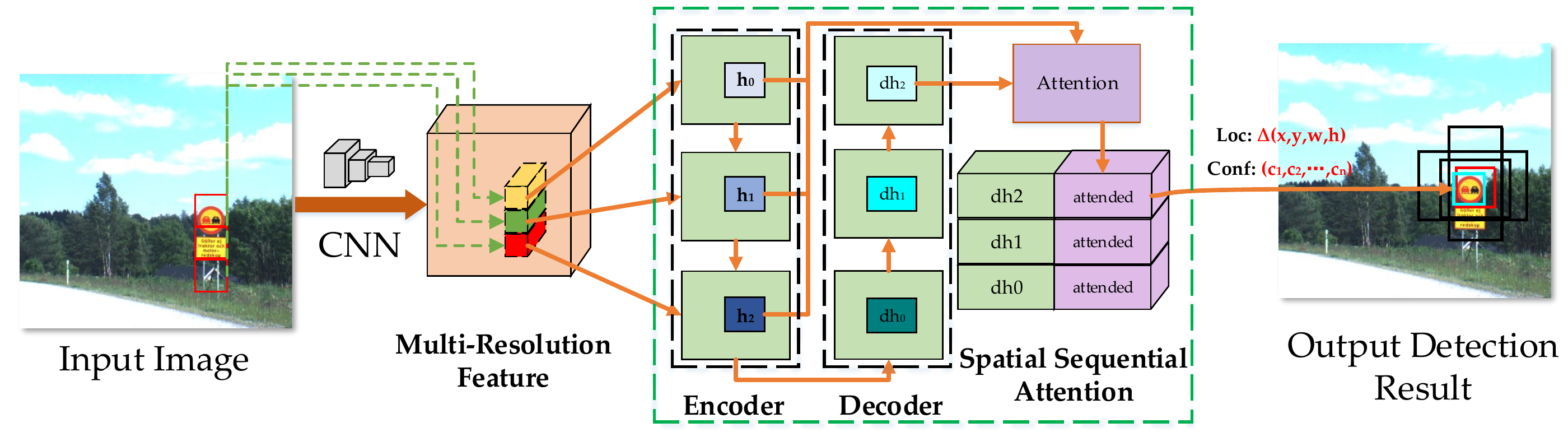}
	\caption{The Vertical spatial sequence attention module.}
	\label{fig:vssa}
\end{figure*}

In addition to the multi-resolution features, context information is also vital for finding small instances, and many previous methods have shown the effectiveness. Unlike existing ones, we treat the traffic sign detection as spatial sequence classification and regression task. The VSSA module is illustrated in Fig. \ref{fig:vssa}. Supposing that an input image is fed into the multi-resolution conv-deconv module, and three feature maps with different resolution are obtained. Then we choose one of the three feature maps as example to describe the VSSA module in detail.

Attention models have been successfully applied in neural machine translation and image captioning tasks. Encoder and decoder models are widely used to model the context for sequence tasks. However, a key limitation of the base encoder-decoder model is that a fixed length context vector is insufficient to encode the sequence. In our model, attention mechanism is used to align the input features at decoding stage and automatically focus on useful features for context modeling.

Based on the chosen feature maps, we divide it into "vertical capsules" to encode the vertical spatial context. For example, a length 3 capsule is a ${(3 \times 1 \times 1024)}$ tensor. We encode the points in one capsule ${(x, y-2),(x, y-1),(x, y)}$ along feature maps of all channels by using LSTM. As shown in Fig. \ref{fig:vssa}, the capsule is comprised of the yellow, green and red feature vectors. We denote the three features from top to bottom at ${\{(x, y-2), (x, y-1), (x, y)\}}$ by ${\{V_{t-2},V_{t-1},V_t\}}$. The LSTM network processes these features step by step, and the context information across the sequences is captured in the hidden state ${h_t}$. This can be formulated as follows: At step t of features from top to bottom, the hidden state is updated with the next input of that sequence by
\begin{equation}
h_t = \mbox{LSTM}(h_{t-1},V_t; \theta),
\end{equation}
We denote these encoded hidden states as ${(h_1,h_2,...,h_T)}$. These three hidden states correspond to features from top to bottom. Moreover, the states are encoded sequentially, so the contextual information is also been encoded from top to bottom. 

Based on the encoded hidden states, another LSTM is used for attention decoding. We denote the decoder hidden states by ${(dh_1,...,dh_T)}$. The attention decoder generate a weight vector over the three encoded hidden states at each decode step ${t}$. By using the attention decoder, every decoding step will exploit features from all three encoded hidden states and generate the next weight vector(attention vector) sequentially. The attention vector ${a_i^t}$ at time ${t}$ can be computed by:

\begin{equation}
p_i^t = v^Ttanh({W_h}h_i+{W_d}dh_{t}),
\end{equation}

\begin{equation}
	a_i^t = softmax(p_i^t),
\end{equation}

Where ${a_i^t}$ is the ${(1 \times T)}$ weight vector generated for the ${t_{th}}$ decoding step. ${v}$, ${W_h}$ and ${W_d}$ are learnable parameters of the model. Then at the decoding stage, we compute the attended hidden state by using the attention vector ${a}$:

\begin{equation}
\hat{dh_t} = \sum_{i=1}^{T}({a_i^t}{h_i}),
\end{equation}

\begin{figure}
	\centering
	\includegraphics[width=0.4\textwidth]{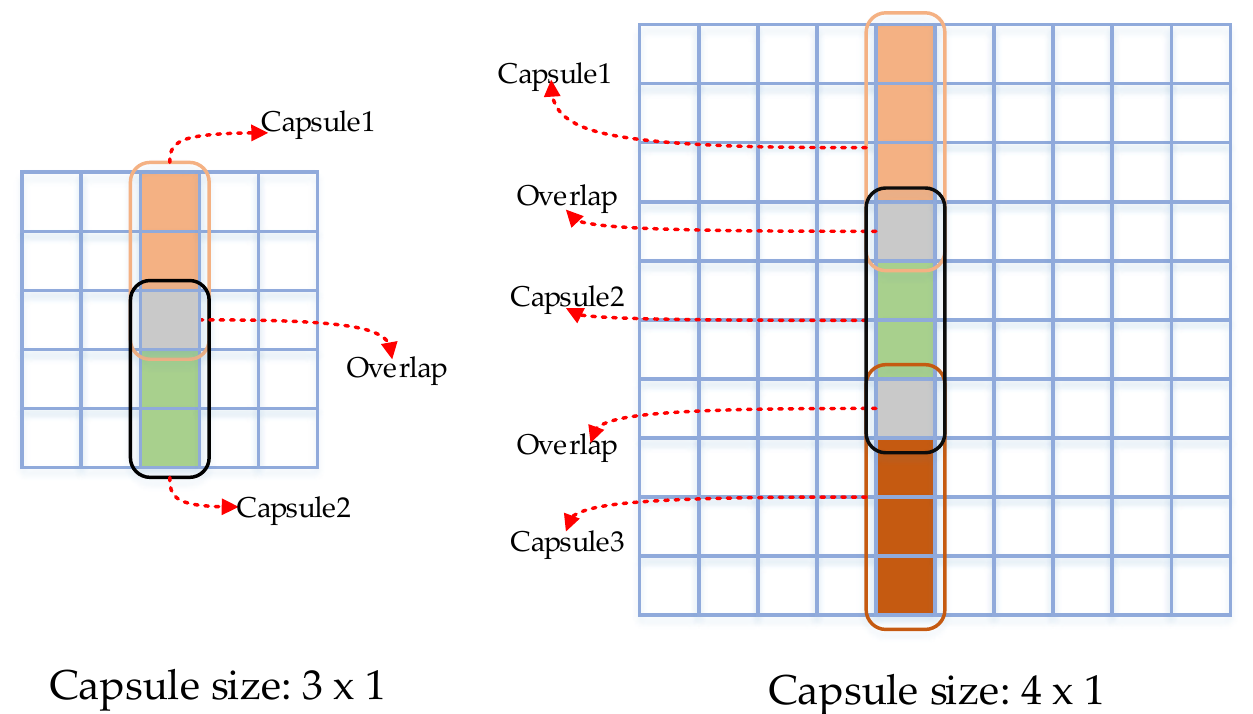}
	\caption{The capsule illustration. Capsules on 5 x 5 and 10 x 10 feature maps are shown.}
	\label{fig:capsule}
\end{figure}
Lastly, we concatenate ${\hat{dh_t}}$ with decoder hidden state ${dh_t}$ together to predict the detection label ${\hat{y}}$ and the bounding box ${\hat{d}}$. It is worth mentioning that we apply the VSSA module on ${DS_{13}}$ and ${DCD_1}$ to reduce the computational complexity. The vertical capsule of the two feature maps are illustrated in Fig. \ref{fig:capsule}. We use capsule size of ${3 \times 1}$ and ${4 \times 1}$ for ${DS_{13}}$ and ${DCD_{1}}$ respectively.

\subsection{Multi-Task Training} Finally, the concatenated ${h_t}$ is convolved by a 1x1 convolution layer and outputs the bounding box shifts together with the classification label for every feature point. If we denote the predicted label and bounding box as ${\hat{y}}$ and ${\hat{d}}$, the total loss function includes the classification log loss ${L_{cls}}$ and the Smooth L1 regression loss:
	\begin{equation}
	\mathcal{L} = \sum_i(L_{cls}(\hat{y_i},y_i) + \alpha{[y>0]}smooth_{L1}(\hat{d_i}-d_i)),
	\end{equation}
where ${\alpha}$ is a fixed parameter to adjust weight between the two loss functions, and we set it to 0.1 in this work. ${y_i}$ is the ground truth label and $d_i$ represents the ground truth bounding box.

\section{Experiments}

\begin{figure}
	\centering
	\includegraphics[width=0.5\textwidth]{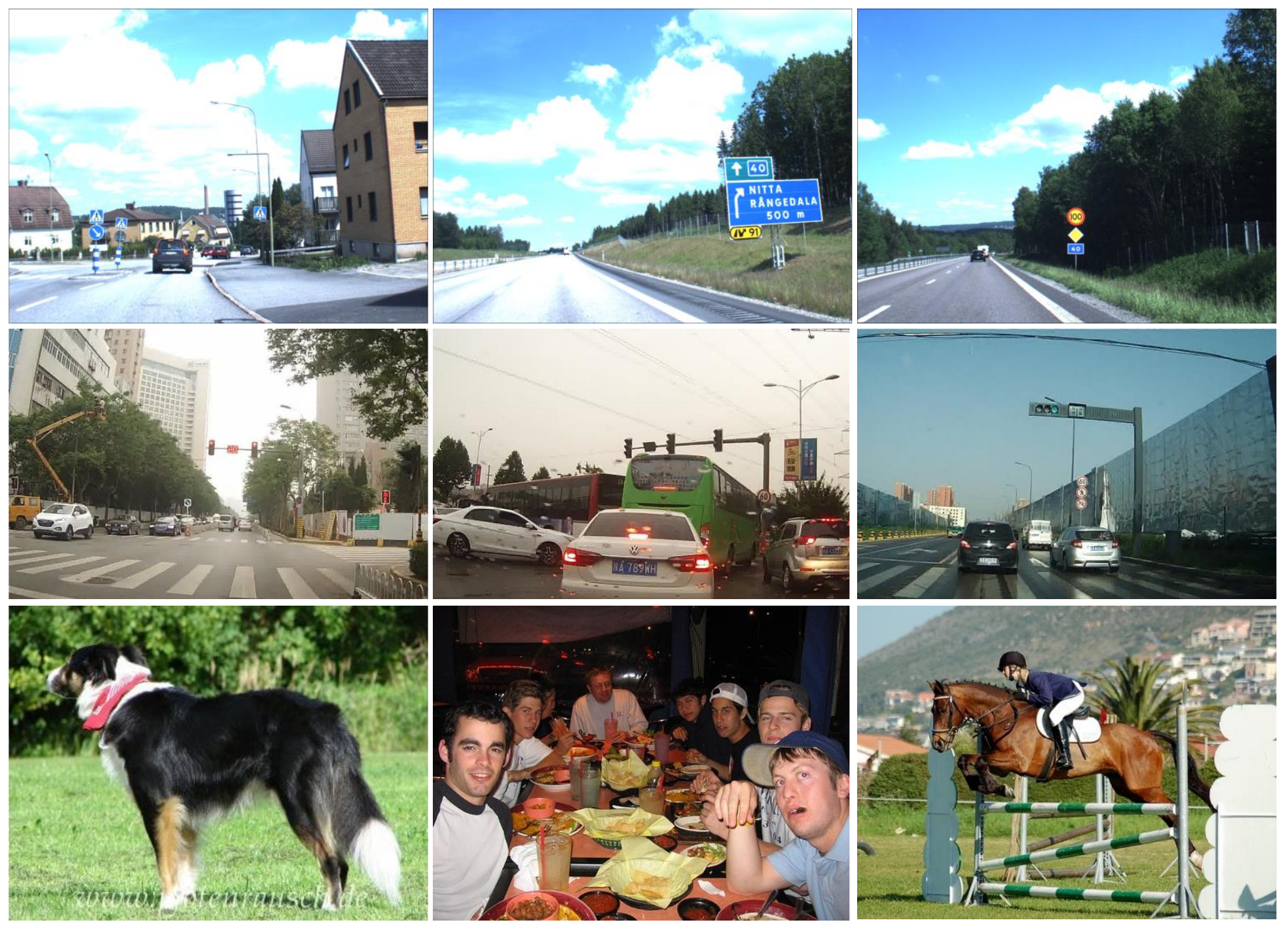}
	\caption{Three datasets used in the experiments. The first row are images in STS dataset. The second row are images from our OPTTSR dataset, and the third row shows images from Pascal VOC 2007 dataset.}
	\label{fig:dd}
\end{figure}

To evaluate the proposed network, we trained and tested it on two traffic sign detection datasets. One is the public STS (Swedish Traffic Signs) dataset \cite{larsson2011using}. STS dataset contains more than 20,000 images and has 20\% labeled for training. There are 3,488 traffic signs captured from highways and cities from more than 350km of Swedish roads in this dataset. Another is our collected traffic sign detection dataset, which contains 8,725 images in total. Images of this dataset are captured in many complicated driving scenes by a camera mounted on the vehicle. We use 5,816 images for training and 2,909 images for evaluation. Additionally, to evaluate the effect of the proposed Multi-Resolution and VSSA module more comprehensively, we conduct experiments on the general object detection dataset, i.e., the Pascal VOC 2007 and 2012 datasets. The image examples are presented in Fig. \ref{fig:dd}. As we can see that in the STS dataset, the traffic scenes are less complex than OPTTSR dataset. Target Objects usually occupy more than 50\% in Pascal voc dataset. In this section, we will describe the parameters setup and the details of the experiments on three datasets.

\subsection{Parameters Setup}
Tensorflow object detection API \cite{huang2017speed} is used to implement the proposed method. We used the MobileNet pretrained model on coco dataset \cite{lin2014microsoft} to initialize the proposed network. The experimental environment is equipped with Intel Core i7 7700K @ 3.50GHz CPU, 64GB RAM, and four NVIDIA GeForce Titan X Pascal GPUs. As for the trained configuration, Stochastic Gradient Descent (SGD) is used to train the model with the following parameters: base input image size 300x300 (for SSD300) or 500x500 (for SSD500), batch size 24, base learning rate 0.0003, initial momentum 0.9, and weight decay 0.0005. 

As the proposed method is based on the single-stage detector SSD, the final multiple feature maps with different resolutions are used to output the detection result. In our experiment setting, 5$\times$5, 10$\times$10 and 19$\times$19 are utilized for detection. All the detection results are processed by the NMS module for the final detection output. For all the SSD based methods, we set the aspect ratios to ${\{1,2,3,\frac{1}{2}, \frac{1}{3}\}}$. For multi-scale training, ${\{0.75, 1, 1.25\}}$ times of the original scale are used.

\begin{table}
	\centering
	\caption{Performance comparison on STS dataset.}
	\begin{tabular}{c|cc}
		\whline
		Methods 								 &Precision(\%)		   &Recall(\%)		\\
		\hline
		\textbf{MRFeature+VSSA(Ours)}	 &\textbf{99.18}   &\textbf{94.42}	\\
		\textbf{MRFeature(Ours)}		 &98.83             &93.96	\\
		FCN \cite{zhu2016traffic} 		   		 &{98.67} &93.27 		\\
		Overfeat \cite{sermanet2013overfeat}                    &95.05				&83.74    \\
		Adaboost SVR \cite{chen2015accurate} 	 &94.52			    &80.85	\\
		R-CNN \cite{girshick2014rich} 	 &90.08 		    &87.27 			\\
		Fourier spatial model \cite{larsson2011using}  &91.84			    &77.08			\\
		\whline
	\end{tabular}\label{table_all}
\end{table}

\begin{table*}[]
	\centering
	\caption{Performance comparison on OPTTSR dataset}
	\scalebox{0.88}{
	\label{tab:opttsr}
	\begin{tabular}{c|c|c|ccccccc|c}
		Method & dataset & mAP & construction & indication & information\_guide & prohibitory & supplemental & tourist & warning & \multicolumn{1}{l}{FPS} \\ \hline
		Faster RCNN + VGG & OPTTSR & 52.00 & 37.40 & 50.83 & 67.35 & 44.77 & 35.77 & 93.46 & 34.47 & 7 \\
		SSD300 + MobileNet & OPTTSR & 48.54 & 34.13 & 45.69 & 60.53 & 39.44 & 36.47 & 92.11 & 31.38 & 46 \\
		SSD500 + MobileNet & OPTTSR & 51.87 & 37.33 & 47.79 & 68.06 & 45.41 & 32.42 & 95.56 & 36.49 & 20 \\
		DSSD321 + Residual-101 & OPTTSR & 53.79 & 37.68 & 48.00 & 68.31 & 45.83 & 38.85 & 96.11 & 41.77 & 9.8 \\ \hline
		SSD300+MRFeature+VSSA(Horizontal) & OPTTSR & 53.45 & 37.37 & 45.14 & 62.29 & 44.25 & 46.86 & 94.68 & 43.56 & 21 \\
		SSD300+MRFeature+VSSA(Vertical) & OPTTSR & 55.43 & 40.48 & 44.49 & 63.60 & 44.85 & \textbf{52.36} & 95.47 & 46.76 & 21 \\
		SSD500+MRFeature+VSSA(Vertical) & OPTTSR & \textbf{59.81} & \textbf{48.55} & \textbf{50.87} & \textbf{73.97} & \textbf{50.90} & 48.09 & \textbf{97.22} & \textbf{49.07} & 8.9
\end{tabular}}
\end{table*}

\subsection{STS dataset}
The detection results on STS dataset is shown in Table \ref{table_all}. The proposed method achieves an average precision of 99.18\% and 94.42\% for average recall on all traffic sign classes, which outperforms other methods and attains state-of-the-art result.  To compare with other traffic sign detection methods, Part0 of Set1 are used as the training set and Part0 of Set2 as the test set. IoU (intersection over union) threshold is set to 0.5 for the evaluation. Following the compared methods, only traffic signs larger than 50 pixels are considered for training and evaluation. SSD with input size ${300 \times 300}$ is used as the base detection framework.

\begin{figure*}
	\centering
	\includegraphics[width=1.0\textwidth]{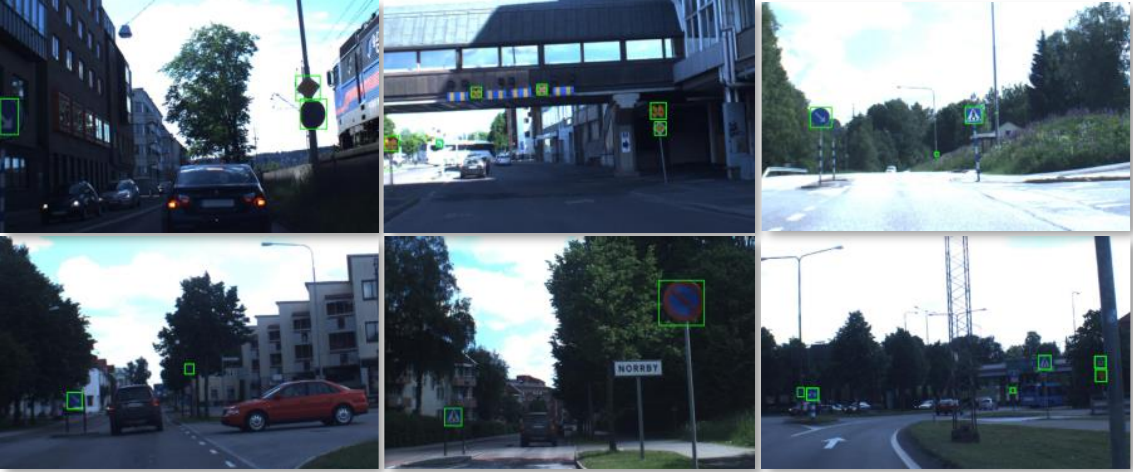}
	\caption{Illustration of some qualitative detection examples tested on STS dataset.}
	\label{fig:stsre}
\end{figure*}

The methods listed in Table \ref{table_all} contain deep learning methods and also conventional methods like \cite{larsson2011using} and \cite{chen2015accurate}. From look on the whole, deep learning methods can obtain better detection performance thanks to the powerful features. To evaluate the effect of the proposed network architecture and the sequence modeling, we trained and tested our method with and without the sequence model separately. From the results we can see that the Multi-Resolution is effective for traffic signs detection even when the sequence model is not used. Moreover, the proposed method with sequence modeling achieves the best detection performance, which indicates the active effect of the proposed context model. Some qualitative detection results on STS dataset is shown in Fig.\ref{fig:stsre}. Although the model is trained with signs whose size is larger than 50 pixel, our method can still detect small traffic signs thanks to the effective Multi-Resolution features and scale-invariant training.

\subsection{OPTTSR dataset}
\begin{figure*}
	\centering
	\includegraphics[width=1.0\textwidth]{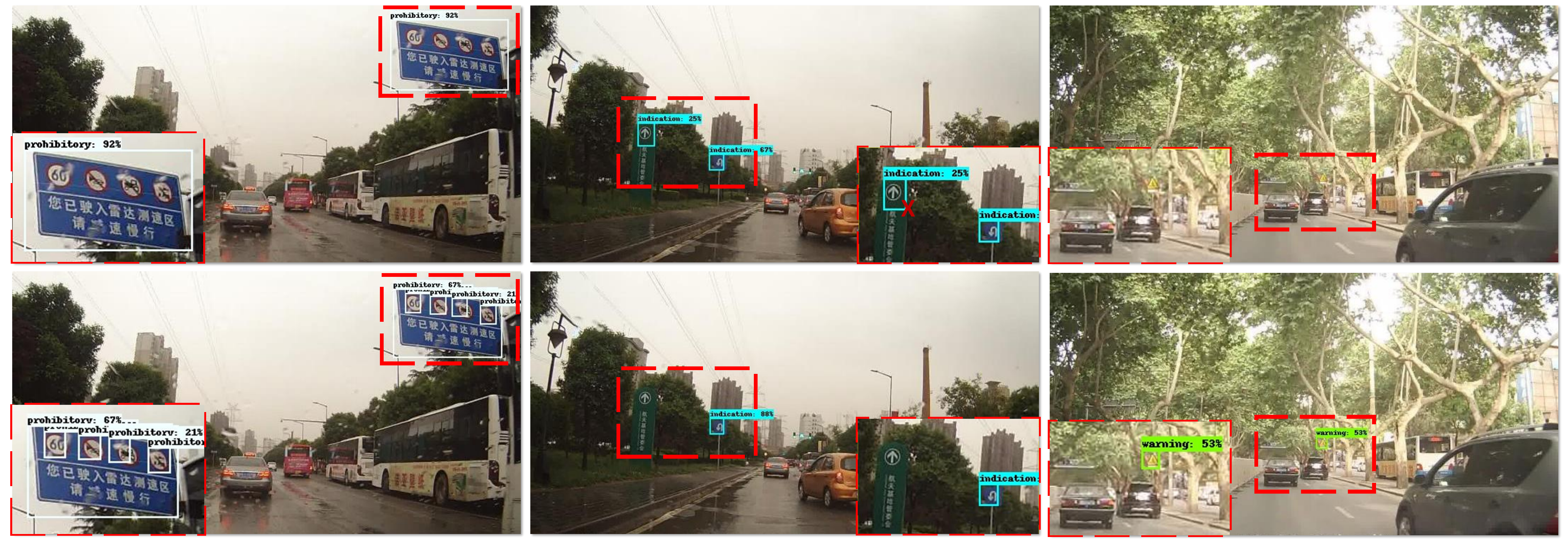}
	\caption{Some detection examples on OPTTSR dataset. The left column shows that the proposed method is more effective for dense signs detection. The middle one and the right one shows our method is more robust than the baseline.}
	\label{fig:opttsr1}
\end{figure*}

\begin{figure*}
	\centering
	\includegraphics[width=1.0\textwidth]{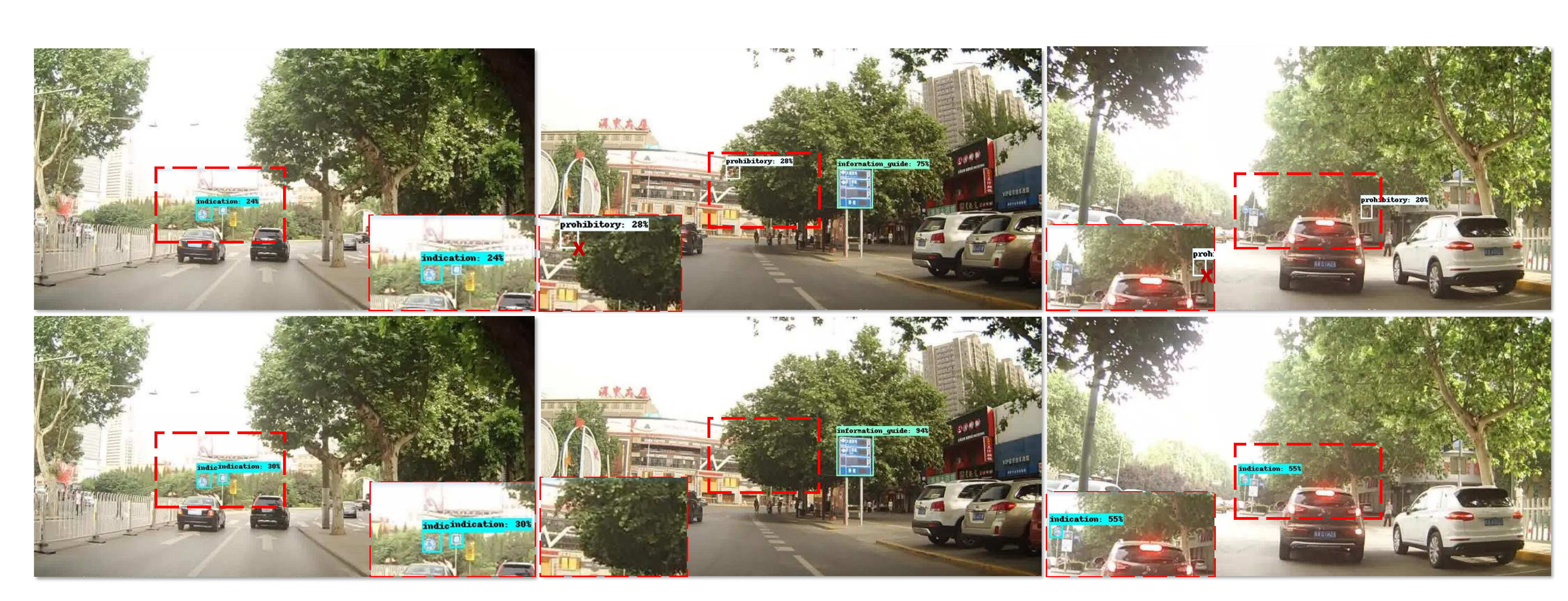}
	\caption{Some qualitative detection examples on OPTTSR dataset.}
	\label{fig:opttsr2}
\end{figure*}

We also perform evaluation on our own traffic sign dataset named OPTTSR, and the detection examples are presented by Fig.\ref{fig:opttsr1} and Fig.\ref{fig:opttsr2}. Our dataset contains 8725 fully annotated images under different and challenge situations. There are 7 class of Chinese traffic signs including construction, indication, information and guide, supplemental, tourist, and warning. In this experiment, SSD with input size ${300 \times 300}$ and ${500 \times 500}$ are used as the base detection framework. The OPTTSR dataset has the same format with Pascal voc 2007 dataset, and mean average precision(mAP) is used as the evaluation measurement.

\begin{figure}
	\centering
	\includegraphics[width=0.48\textwidth]{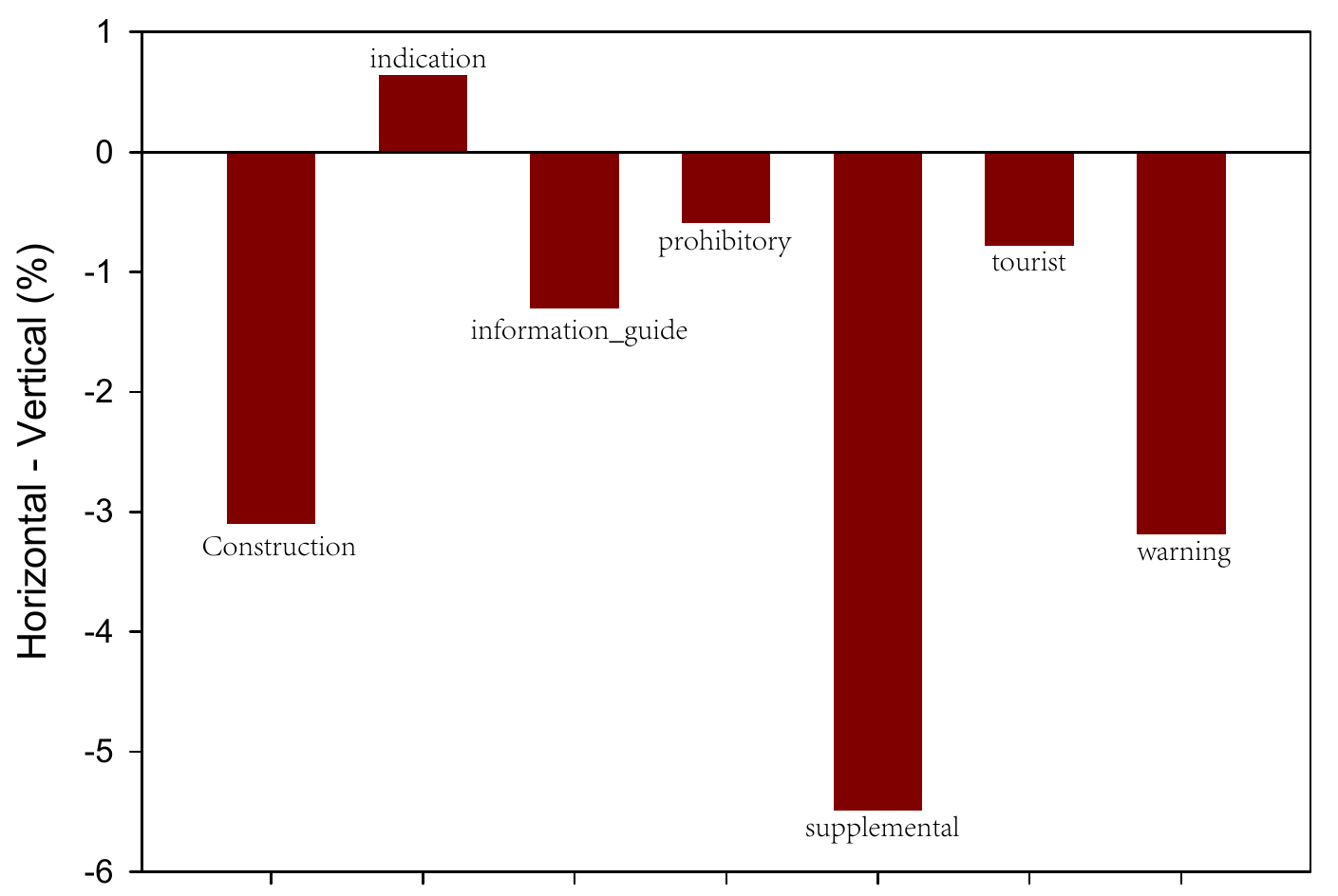}
	\caption{The comparison of the horizontal and vertical direction context modeling. We use "Horizontal - Vertical" for a clearly presentation of the difference on OPTTSR dataset.}
	\label{fig:ajb}
\end{figure}

The detail detection results on OPTTSR dataset are shown in Table \ref{tab:opttsr}. It is worth mentioning that the size of annotated signs in OPTTSR dataset ranges from 10 pixel to 400 pixel, and there are many challenging situations such as rainy weathers, illumination variation and serious occlusion. To evaluate the proposed method, we compared 4 existing popular detection architectures, i.e., Faster RCNN with VGG16 network, SSD 300 with MobileNet network, SSD 500 with MobileNet network and Deconvolutional SSD with 321 input size and ResNet 101 as backbone. Moreover, to analyze the effect of different spatial direction context, we conduct experiments to compare the models with horizontal and vertical spatial context. The results are presented in Tab. \ref{tab:opttsr}. From the results we can see that our method with vertical spatial modeling outperforms it with horizontal direction. There are obvious performance improvement for the construction" and "supplemental" signs. In Fig. \ref{fig:ajb}, there is a more clear display to show the difference of the two models. For 6 out of 7 types of signs, the "vertical" spatial sequence model performs better than the horizontal one. This result supports the assumption that vertical context is more important and effective than the horizontal information for traffic sign detection task.

The baseline detection method merely achieve mAP 48.54\%. However, with the proposed two modules, a mAP of 55.43\% can be obtained, which is a significant improvement compared to the baseline method. Moreover, we observe that the improvement is mainly caused by the better performance for small size traffic sign. The reason may be that in complex traffic scenes, there are many objects which have similar appearance with traffic signs such as billboards or other signs unrelated to traffic. The proposed VSSA module can help to reduce this kind of mistakes by encoding more context information. From the results in Table. \ref{tab:opttsr} we can see that the best detection result is obtained by the proposed method with explicit vertical spatial context encoding. 

As real-time processing is a required ingredient for application like traffic sign detection task, we evaluate and compare the inference time of the proposed method as well as other methods in Tab. \ref{tab:opttsr}. Although single-stage based detection frameworks achieve slightly lower performance than Faster RCNN, they substantially outperform the two-stage based methods in terms of time-efficiency. As we can see, "SSD 500" outperforms the method "SSD 300" by using larger input scale. However, with the multi-resolution feature learning and context modeling, the proposed method can improve the "SSD 300" baseline by a significant gap of 6.89\%. The proposed vertical context modeling method with 300$\times$300 input can operate at 21 frames per second (FPS), and it can still obtain better performance than the "DSSD 321" method with a running speed of 9.8 FPS. To explore the effect of the input size, we used the 500$\times$500 input scale for the proposed method, and `SSD500+MRFeature+VSSA(Vertical)' can further improve the performance at a cost of the running time.

Additionally, capsule size is another critical factor which affect the inference time of the proposed method. In terms of performance, optimal capsule size is hard to find. But in our experiments, we observe that capsule sizes such as 3, 4 and 5 performs well on our dataset and larger capsule size will dramatically increase the inference time. Considering this, capsules are only used on some of the multi-resolution feature maps, i.e., the 10$\times$10 and 5$\times$5 maps in our experiments. Another reason for choosing small capsule size on high-level feature maps is that one pixel on these feature maps corresponds to a large receptive field on the original image. Thus too large capsule size is not considered in our experiments.

Qualitative detection results of OPTTSR dataset is shown in Fig. \ref{fig:opttsr1} and Fig. \ref{fig:opttsr2}. The first row of Fig. \ref{fig:opttsr1} and Fig. \ref{fig:opttsr2} are the results of SSD500 + MobileNet, i.e., the baseline detection method, and the second row are the results of the proposed SSD500+MRFeature+VSSA method. Our proposed method can reduce false positive detections and get better detection performance.

\begin{table*}[]
	\centering
	\caption{Comparison Results on Pascal voc 2007 dataset}
	\scalebox{0.7}{
		\label{tabel:voc}
		\begin{tabular}{c|c|c|cccccccccccccccccccc}
			Method                & dataset & mAP           & aero          & bike          & bird & boat          & bottle & bus           & car           & cat           & chair & cow           & table & dog           & horse         & mbike         & person        & plant & sheep         & sofa          & train         & tv            \\ \hline
			Fast RCNN             & voc07   & 66.9          & 74.5          & 78.3          & 69.2                                 & 53.2          & 36.6                                   & 77.3          & 78.2          & 82.0          & 40.7                                  & 72.7          & 67.9                                  & 79.6          & 79.2          & 73.0          & 69.0          & 30.1                                  & 65.4          & 70.2          & 75.8          & 65.8          \\
			Faster RCNN           & voc07   & 69.9          & 70.0          & 80.6          & 70.1                                 & 57.3          & 49.9                                   & 78.2          & 80.4          & 82.0          & 52.2                                  & 75.3          & 67.2                                  & 80.3          & 79.8          & 75.0          & 76.3          & 39.1                                  & 68.3          & 67.3          & 81.1          & 67.6          \\
			SSD300+VGG            & voc07   & 68.0          & 73.4          & 77.5          & 64.1                                 & 59.0          & 38.9                                   & 75.2          & 80.8          & 78.5          & 46.0                                  & 67.8          & 69.2                                  & 76.6          & 82.1          & 77.0          & 72.5          & 41.2                                  & 64.2          & 69.1          & 78.0          & 68.5          \\ \hline
			SSD300+MobileNet      & voc07   & 71.3          & 73.1          & 81.3          & 68.2                                 & 62.0          & 38.8                                   & 84.6          & 83.6          & 83.9          & 49.3                                  & 70.9          & 72.4                                  & 80.6          & \textbf{85.5} & 80.8          & 77.7          & 40.8                                  & \textbf{71.8}          & \textbf{74.7} & 86.3          & 71.5          \\
			SSD300+MRFeature      & voc07   & 72.6          & \textbf{79.8} & 80.2          & 70.2                                 & \textbf{63.5} & \textbf{41.3}                          & 83.6          & \textbf{83.8} & \textbf{86.3} & 51.3                                  & \textbf{74.7} & 70.7                                  & 81.3          & 81.6          & 80.1          & 79.0          & 43.2                                  & 67.6          & 73.0          & \textbf{87.7} & 72.5          \\
			SSD300+MRFeature+VSSA & voc07   & \textbf{73.4} & 78.0          & \textbf{82.3} & \textbf{70.8}                        & 62.8          & 40.2                                   & \textbf{85.2} & 83.7          & 85.7          & \textbf{52.4}                         & 74.0          & \textbf{74.9}                         & \textbf{82.9} & 84.7          & \textbf{81.8} & \textbf{79.5} & \textbf{45.7}                         & 69.7 & 73.1          & 87.0          & \textbf{73.1}
	\end{tabular}}
\end{table*}

\begin{table*}[]
	\centering
	\caption{Results on Pascal voc 2007 dataset}
	\scalebox{0.7}{
		\label{tabel:voc12}
		\begin{tabular}{c|c|c|cccccccccccccccccccc}
			Method                  & dataset                     & mAP           & aero          & bike          & bird          & boat          & bottle        & bus           & car           & cat           & chair         & cow           & table         & dog           & horse         & mbike         & person        & plant         & sheep         & sofa          & train         & tv            \\ \hline
			Faster RCNN             & 07++12                      & 73.2          & 76.5          & 79.0          & 70.9          & 65.5          & 52.1          & 83.1          & 84.7          & 86.4          & 52.0          & 81.9          & 65.7          & 84.8          & 84.6          & 77.5          & 76.7          & 38.8          & 73.6          & 73.9          & 83.0          & 72.6          \\
			YOLOv2 544              & 07++12                      & 73.4          & 86.3          & 82.0          & 74.8          & 59.2          & 51.8          & 79.8          & 76.5          & 90.6          & 52.1          & 78.2          & 58.5          & 89.3          & 82.5          & 83.4          & 81.3          & 49.1          & 77.2          & 62.4          & 83.8          & 68.7          \\
			SSD300+VGG              & 07++12                      & 75.8          & 88.1          & 82.9          & 74.4          & 61.9          & 47.6          & 82.7          & 78.8          & 91.5          & 58.1          & 80.0          & 64.1          & 89.4          & 85.7          & 85.5          & 82.6          & 50.2          & 79.8          & 73.6          & 86.6          & 72.1          \\
			DSSD321+Residual-101    & 07++12                      & 76.3          & 87.3          & 83.3          & 75.4          & 64.6          & 46.8          & 82.7          & 76.5          & 92.9          & 59.5          & 78.3          & 64.3          & 91.5          & 86.6          & 86.6          & 82.1          & 53.3          & 79.6          & 75.7          & 85.2          & 73.9          \\
			DSSD513+Residual-101    & 07++12                      & \textbf{80.0} & \textbf{92.1} & 86.6          & 80.3          & 68.7          & 58.2          & \textbf{84.3} & 85.0          & \textbf{94.6} & 63.3          & \textbf{85.9} & 65.6          & \textbf{93.0} & 88.5          & \textbf{87.8} & \textbf{86.4} & \textbf{57.4} & \textbf{85.2} & 73.4          & 87.8          & \textbf{76.8} \\
			Perceptual GAN + VGG    & \multicolumn{1}{l|}{07++12} & -             & -             & -             & -             & 69.4          & \textbf{60.2} & -             & -             & -             & 57.9          & -             & -             & -             & -             & -             & -             & 41.8          & -             & -             & -             & -             \\ \hline
			SSD300+VSSA(Horizontal) & 07++12                      & 78.5          & 80            & 88.3          & \textbf{83.2} & 69.7          & 53.9          & 82.8          & 86.7          & 86.4          & 64.1          & 83.7          & 74.2          & 89.4          & \textbf{90.7} & 84.6          & 76.4          & 50.7          & 80.2          & \textbf{83.5} & 88            & 74            \\
			SSD300+VSSA(Vertical)   & 07++12                      & 78.7          & 81.5          & \textbf{88.4} & 82.7          & \textbf{72.8} & 55.4          & 83.5          & \textbf{87.5} & 87.6          & \textbf{65.2} & 83.1          & \textbf{74.5} & 86.3          & 90.0          & 83.4          & 75.8          & 50.1          & 80.8          & 82.4          & \textbf{88.9} & 73.1         
	\end{tabular}}
\end{table*}

\begin{figure}
	\centering
	\includegraphics[width=0.48\textwidth]{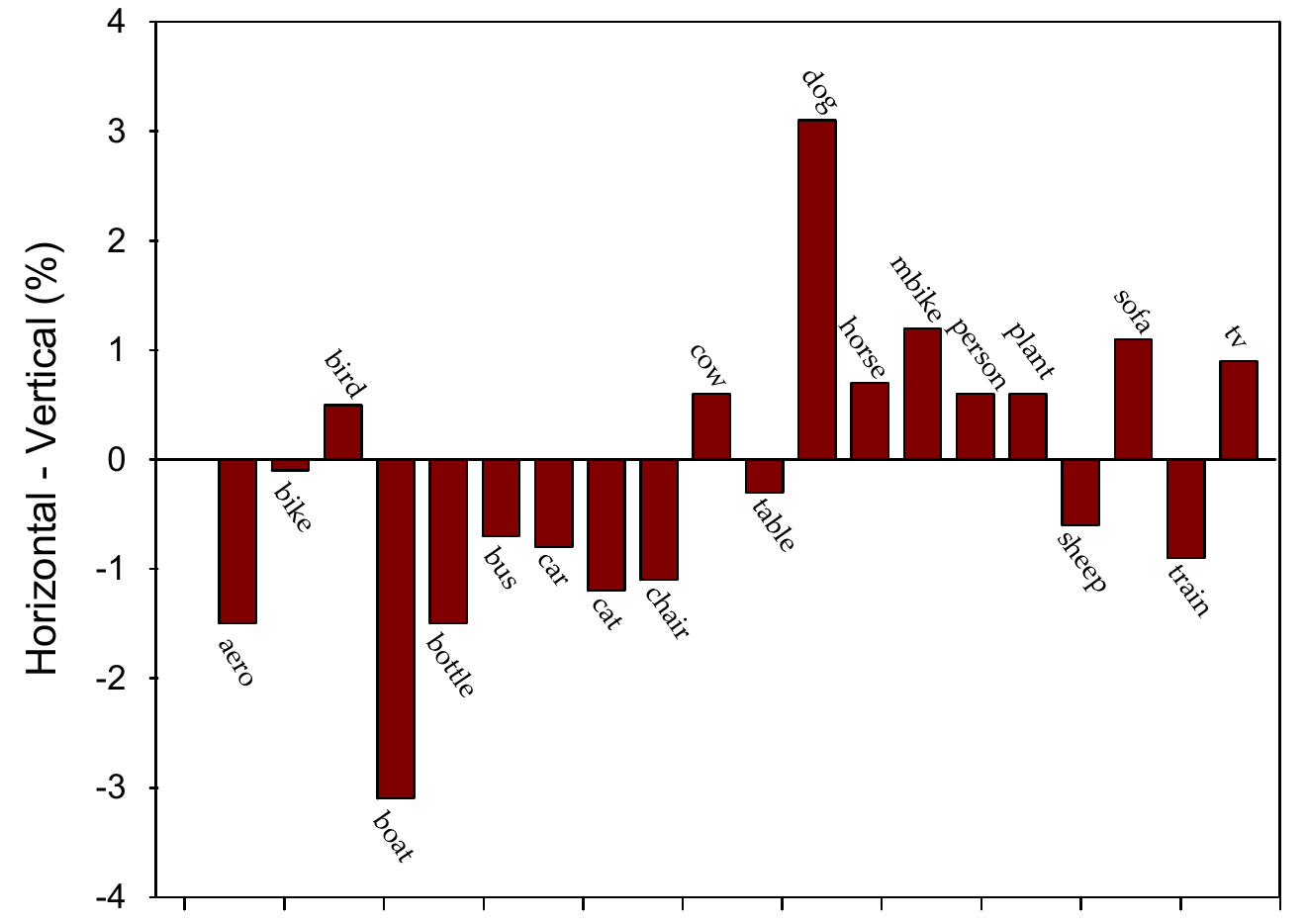}
	\caption{The comparison of the horizontal and vertical direction context modeling. We use "Horizontal - Vertical" for a clearly presentation of the difference on Pascal VOC 2007 dataset.}
	\label{fig:amb}
\end{figure}

\begin{figure}
	\centering
	\includegraphics[width=0.5\textwidth]{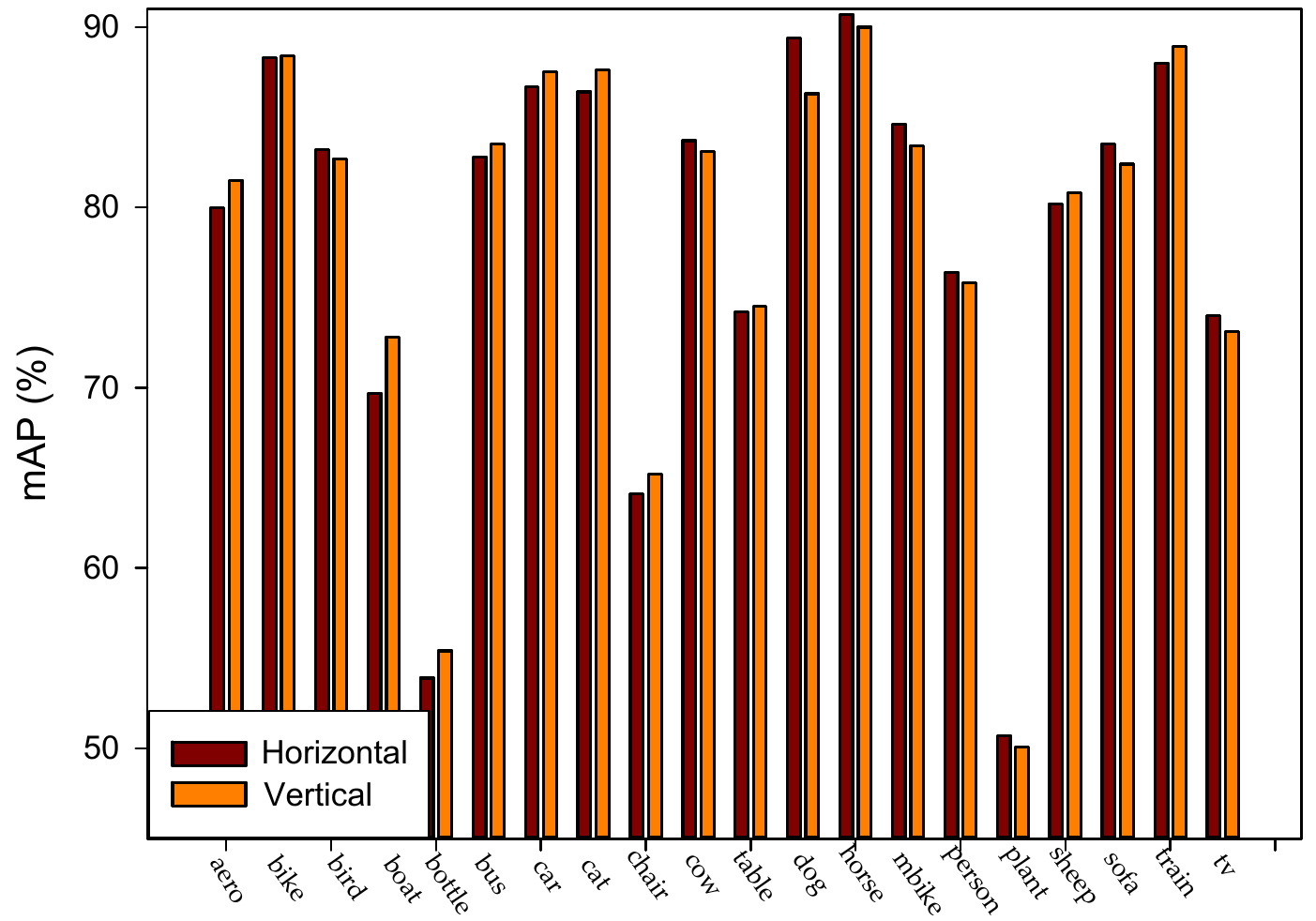}
	\caption{The comparison results of vertical and horizontal directional context models on Pascal VOC 2007 dataset.}
	\label{bar_graph}
\end{figure}

\subsection{Pascal voc dataset}
To get a more comprehensive evaluation of the proposed method, we test it on Pascal VOC 2007, which is a 20 class general object detection dataset. 5011 images are used for training and 4952 images are used for evaluation. The detection performance comparison is presented in Table. \ref{tabel:voc}. We trained the SSD based detection methods for 70000 iterations. The experiment results indicate that the full proposed method achieves the best detection performance.

\begin{figure*}
	\centering
	\includegraphics[width=1.0\textwidth]{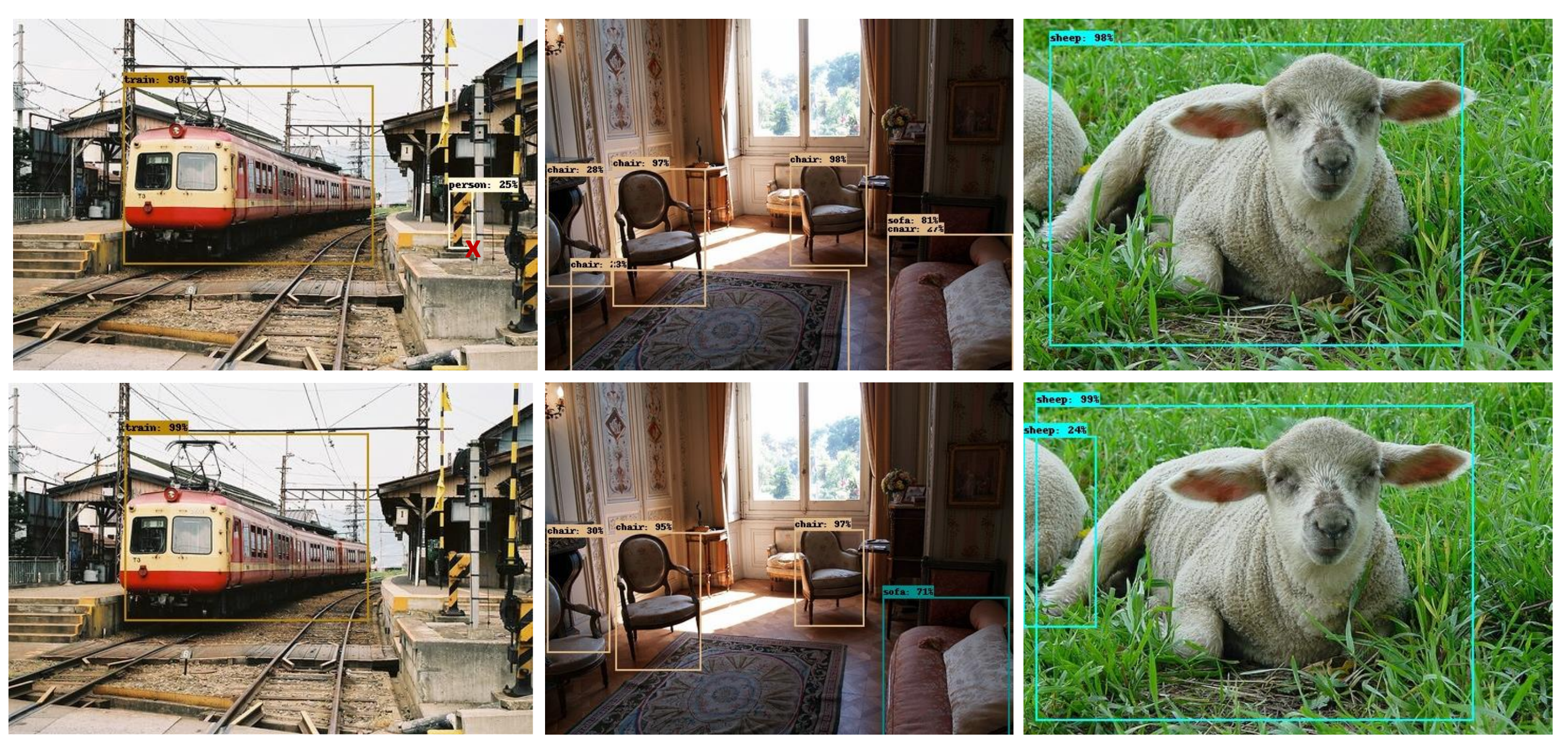}
	\caption{Some qualitative detection examples on Pascal VOC07 dataset.}
	\label{fig:pascal1}
\end{figure*}

\begin{figure*}
	\centering
	\includegraphics[width=1.0\textwidth]{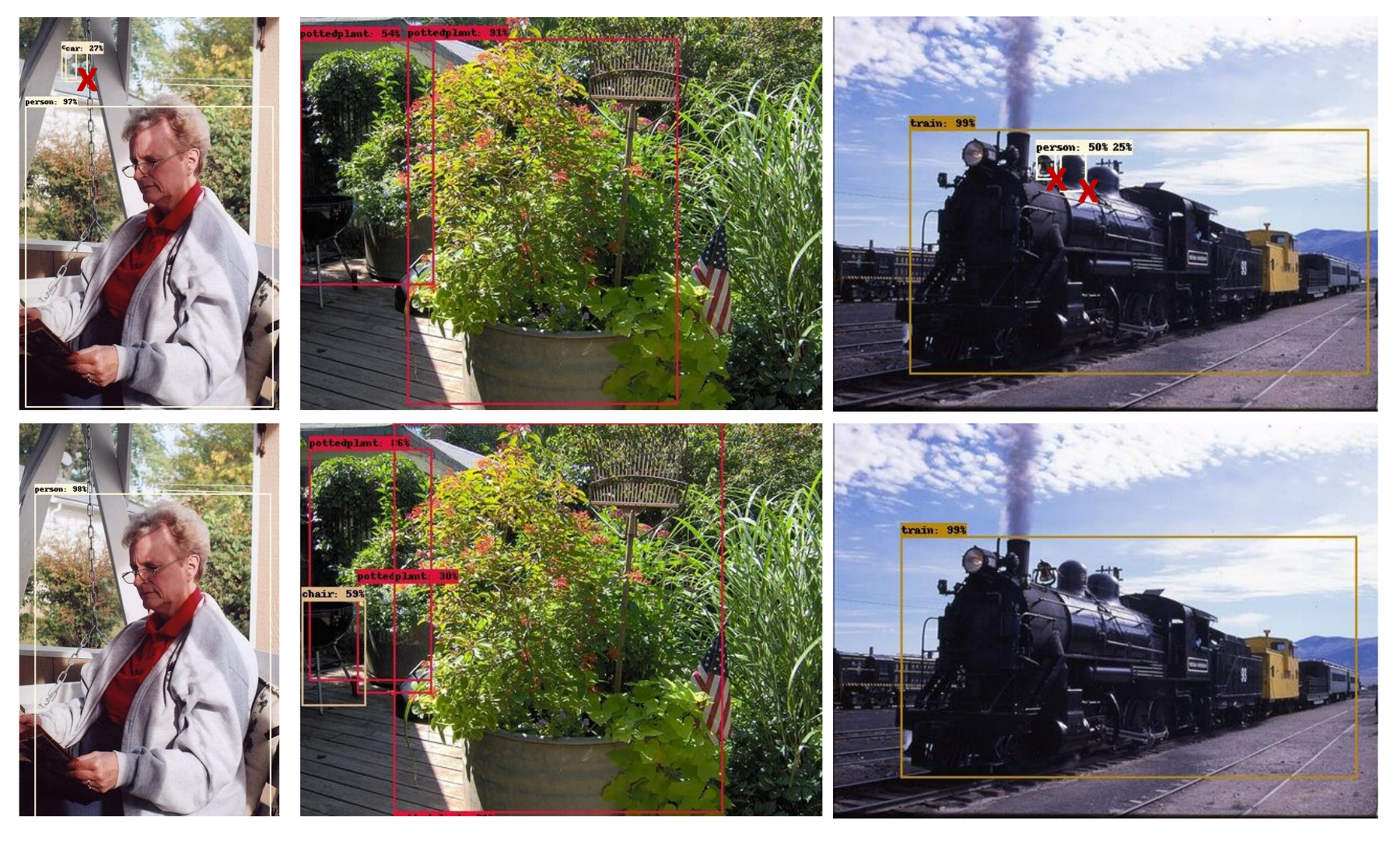}
	\caption{Some qualitative detection examples on Pascal VOC07 dataset.}
	\label{fig:pascal2}
\end{figure*}

As we can see from Table \ref{tabel:voc}, the proposed method achieves better detection performance than the baseline methods. However, the 2.1\% improvement is mainly caused by several kinds of objects. Taking the "chair" class for example, as the chairs are put on the ground with a natural vertical direction, so the vertical spatial context is more effective for this kind of object. Some qualitative detection results are presented in Fig. \ref{fig:pascal1} and Fig. \ref{fig:pascal2}. the first row in the result of the baseline detection method and the second row is the result of the full proposed method. These figures indicated that the proposed method can achieve better detection performance than other methods especially for objects with vertical spatial property.

Additionally, we compare the proposed horizontal and vertical "SSD300+VSSA" methods with other 6 state-of-the-art methods on Pascal VOC 2007 with Pascal VOC 2012 augmented dataset, and the results are shown in Tab. \ref{tabel:voc12}. Among these methods, DSSD is the most relevant method to the proposed one considering that the deconvolution layer is used to boost the features. Perceptual GAN \cite{li2017perceptual} employs generator to transfer the features of small object to super-resolved one. It deals with the small object detection problem by obtaining the super-resolved feature. The results reveal that the proposed method with context modeling outperform the baseline "SSD300+VGG" and the DSSD321 methods. Although DSSD 513 with ResNet 101 network outperforms our method, it is more time-consuming with large input size ${513\times 513}$ and ResNet 101 as backbone.

It is worth mentioning that models with the two direction context achieve similar mAP on the general object detection dataset, which is different from the traffic sign dataset. Most interestingly, models with different direction context can improve the detection performance of several specific object types. From Fig. \ref{fig:amb} we can see that vertical direction context modeling improves the performance of "boat" and "bottle" more obviously. However, horizontal context modeling do good to the "dog" and "sofa" objects. To clearly compare different spatial-direction models, we illustrate the results of them in Fig. \ref{bar_graph}.

\section{Conclusion}
In this work, we treat the traffic sign detection task as a sequence classification and regression task, and proposed a unified end-to-end traffic sign detection framework. The main network consists of two modules. One is the Multi-Resolution feature learning module and another is the vertical spatial sequence attention (VSSA) module. The Multi-Resolution module is constructed by concatenating multi-layer features to densely connected deconvolution layers. Through the Multi-Resolution module, multiple resolution feature maps with higher semantic level are obtained. In addition, context information is taken into consideration by using sequence classification and regression with attention mechanism. From the experiments we can see that many false positive detentions can be suppressed by this module. Finally, the proposed network is evaluated on STS dataset, our OPTTSR dataset and Pascal VOC 2007 dataset. The results have shown the effectiveness of the proposed method.

\section{Acknowledgment}
This work was supported in part by the National Key R\&D Program of China under Grant 2017YFB1002202, in part by the State Key Program of National Natural Science Foundation of China under Grant 61632018, in part by the National Natural Science Foundation of China under Grant 61773316, in part by the Natural Science Foundation of Shaanxi Province under Grant 2018KJXX-024, in part by the Fundamental Research Funds for the Central Universities under Grant 3102017AX010, in part by the Open Research Fund of Key Laboratory of Spectral Imaging Technology, Chinese Academy of Sciences, and in part by the Projects of Special Zone for National Defense Science and Technology Innovation.

%\vfill
%\pagebreak
\begin{IEEEbiography}[{\includegraphics[width=1in,height=1.25in,clip,keepaspectratio]{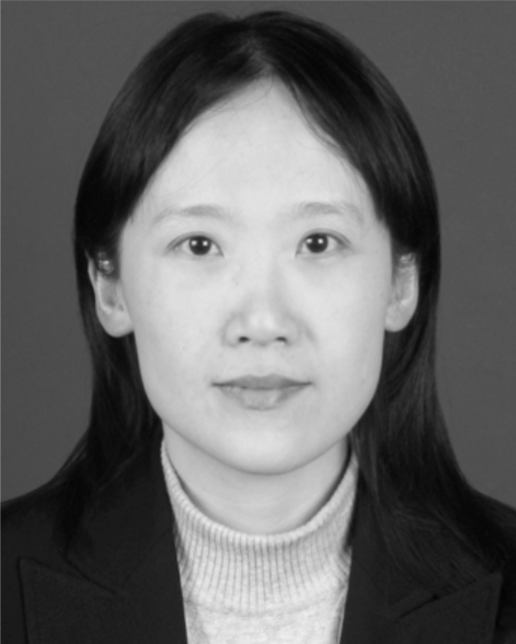}}]{Yuan Yuan} (M'05-SM'09) is currently a Full Professor with the School of Computer Science and the Center for Optical Imagery Analysis and	Learning, Northwestern Polytechnical University, Xi'an, China. She has authored or co-authored over 150 papers, including about 100 in reputable journals, such as the IEEE TRANSACTIONS AND PATTERN RECOGNITION, as well as the conference papers in CVPR, BMVC, ICIP, and ICASSP.	Her current research interests include visual information processing and image/video content analysis.
\end{IEEEbiography}

\begin{IEEEbiography}[{\includegraphics[width=1in,height=1.25in,clip,keepaspectratio]{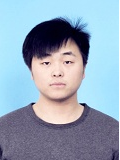}}]{Zhitong Xiong} received the M.E. degree in Northwestern Polytechnical University and is currently working toward the	Ph.D. degree with the School of Computer Science and Center for OPTical IMagery Analysis and Learning (OPTIMAL), Northwestern Polytechnical University, Xi'an, China. His research interests include computer vision and machine learning.
\end{IEEEbiography}

\begin{IEEEbiography}[{\includegraphics[width=1in,height=1.25in,clip,keepaspectratio]{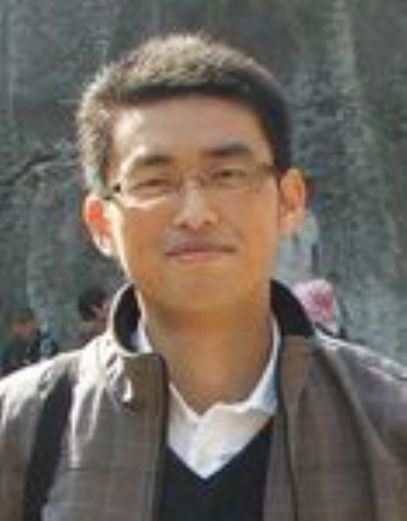}}]{Qi Wang} (M'15-SM'15) received the B.E.	degree in automation and the Ph.D. degree in pattern recognition and intelligent systems from the University of Science and Technology of China,	Hefei, China, in 2005 and 2010, respectively. He is	currently a Professor with the School of Computer Science and the Center for Optical Imagery Analysis and Learning, Northwestern Polytechnical University, Xi'an, China. His research interests include computer vision and pattern recognition.
\end{IEEEbiography}

\bibliographystyle{IEEEbib}
\bibliography{refs}

\end{document}